\documentclass[lettersize,journal]{IEEEtran}
\usepackage{amsmath,amsfonts}
\usepackage{array}
\usepackage[caption=false,font=normalsize,labelfont=sf,textfont=sf]{subfig}
\usepackage{textcomp}
\usepackage{stfloats}
\usepackage{url}
\usepackage{verbatim}
\usepackage{graphicx}
\usepackage{cite}
\usepackage{amsmath}
\usepackage{autobreak}

\usepackage{algorithmic}
\usepackage{booktabs}
\usepackage{bm}
\usepackage{float}
\usepackage{soul}
\usepackage{xcolor}
\usepackage{multirow}
\usepackage{graphicx}
\usepackage[linesnumbered,boxed,ruled,commentsnumbered]{algorithm2e}
\usepackage{textcomp}
\usepackage{threeparttable}
\begin{document}

\title{Progression Cognition Reinforcement Learning with Prioritized Experience for Multi-Vehicle Pursuit}

\author{Xinhang Li,~\IEEEmembership{Graduate Student Member,~IEEE}, Yiying Yang, Zheng Yuan, Zhe Wang,~\IEEEmembership{Graduate Student Member,~IEEE}, Qinwen Wang, Chen Xu,~\IEEEmembership{Member,~IEEE}, Lei Li, Jianhua He,~\IEEEmembership{Senior Member,~IEEE}, and Lin Zhang\IEEEauthorrefmark{1},~\IEEEmembership{Member,~IEEE}
\thanks{\IEEEauthorrefmark{1}Corresponding author}
\thanks{X. Li, Y. Yang, Z. Yuan, Q. Wang, C. Xu and L. Li are with the School of Artificial Intelligence, Beijing University of Posts and Telecommunications, Beijing 100876, China (e-mail: {lixinhang, yyying, yuanzheng, wangqinwen, chen.xu, leili}@bupt.edu.cn). }
\thanks{Z. Wang is with Centre for Telecommunications Research, King’s College London, London WC2R 2LS, U.K. (e-mail: tylor.wang@kcl.ac.uk).}
\thanks{J. He is with School of Computer Science and Electronics Engineering, University of Essex, Colchester, U.K. (e-mail: j.he@essex.ac.uk).}
\thanks{L. Zhang is with the School of Artificial Intelligence, Beijing University of Posts and Telecommunications, Beijing 100876, China, and also with the Beijing Big Data Center, Beijing, China (e-mail: zhanglin@bupt.edu.cn).}

}

\markboth{IEEE Transactions on Intelligent Transportation Systems,~Vol.~XX, No.~X, xxxxx~xxxx}%
{Shell \MakeLowercase{\textit{et al.}}: A Sample Article Using IEEEtran.cls for IEEE Journals}

\IEEEpubid{0000--0000~\copyright~2023 IEEE}

\maketitle

\begin{abstract}
Multi-vehicle pursuit (MVP) such as autonomous police vehicles pursuing suspects is important but very challenging due to its mission and safety critical nature. While multi-agent reinforcement learning (MARL) algorithms have been proposed for MVP problem in structured grid-pattern roads, the existing algorithms use randomly training samples in centralized learning, which leads to homogeneous agents showing low collaboration performance. For the more challenging problem of pursuing multiple evading vehicles, these algorithms typically select a fixed target evading vehicle for pursuing vehicles without considering dynamic traffic situation, which significantly reduces pursuing success rate. To address the above problems, this paper proposes a Progression Cognition Reinforcement Learning with Prioritized Experience for MVP (PEPCRL-MVP) in urban multi-intersection dynamic traffic scenes. PEPCRL-MVP uses a prioritization network to assess the transitions in the global experience replay buffer according to the parameters of each MARL agent. With the personalized and prioritized experience set selected via the prioritization network, diversity is introduced to the learning process of MARL, which can improve collaboration and task related performance. Furthermore, PEPCRL-MVP employs an attention module to extract critical features from complex urban traffic environments. These features are used to develop progression cognition method to adaptively group pursuing vehicles. Each group efficiently target one evading vehicle in dynamic driving environments. Extensive experiments conducted with a simulator over unstructured roads of an urban area show that PEPCRL-MVP is superior to other state-of-the-art methods. Specifically, PEPCRL-MVP improves pursuing efficiency by 3.95$\%$ over TD3-DMAP and its success rate is 34.78$\%$ higher than that of MADDPG. Codes are open sourced.
\end{abstract}


\section{Introduction}
\IEEEPARstart{E}{mpowered} by the self-learning ability of reinforcement learning (RL) and significantly improved environment perception, autonomous driving (AD) \cite{a5,a3,a6} is growing with fast pace and great potentials to improve driving safety and traffic efficiency \cite{a1,a2,a4}. Multi-vehicle pursuit (MVP) is a specific application of AD technology, where multiple autonomous pursuing vehicles chasing one or more suspects vehicles. One example of MVP applications is pursuing criminal suspects by police vehicles, which is specifically illustrated by the New York City Police Department's guidelines \cite{Patrol_guide}. The MVP tasks are usually mission and safety critical. Efficient multi-vehicle collaboration and comprehensive perception under complex and dynamic traffic environments are important to successfully complete the MVP tasks \cite{mvp3}.

\begin{figure*}[]
\centering
\centerline{\includegraphics[width=\textwidth,height=10.6cm]{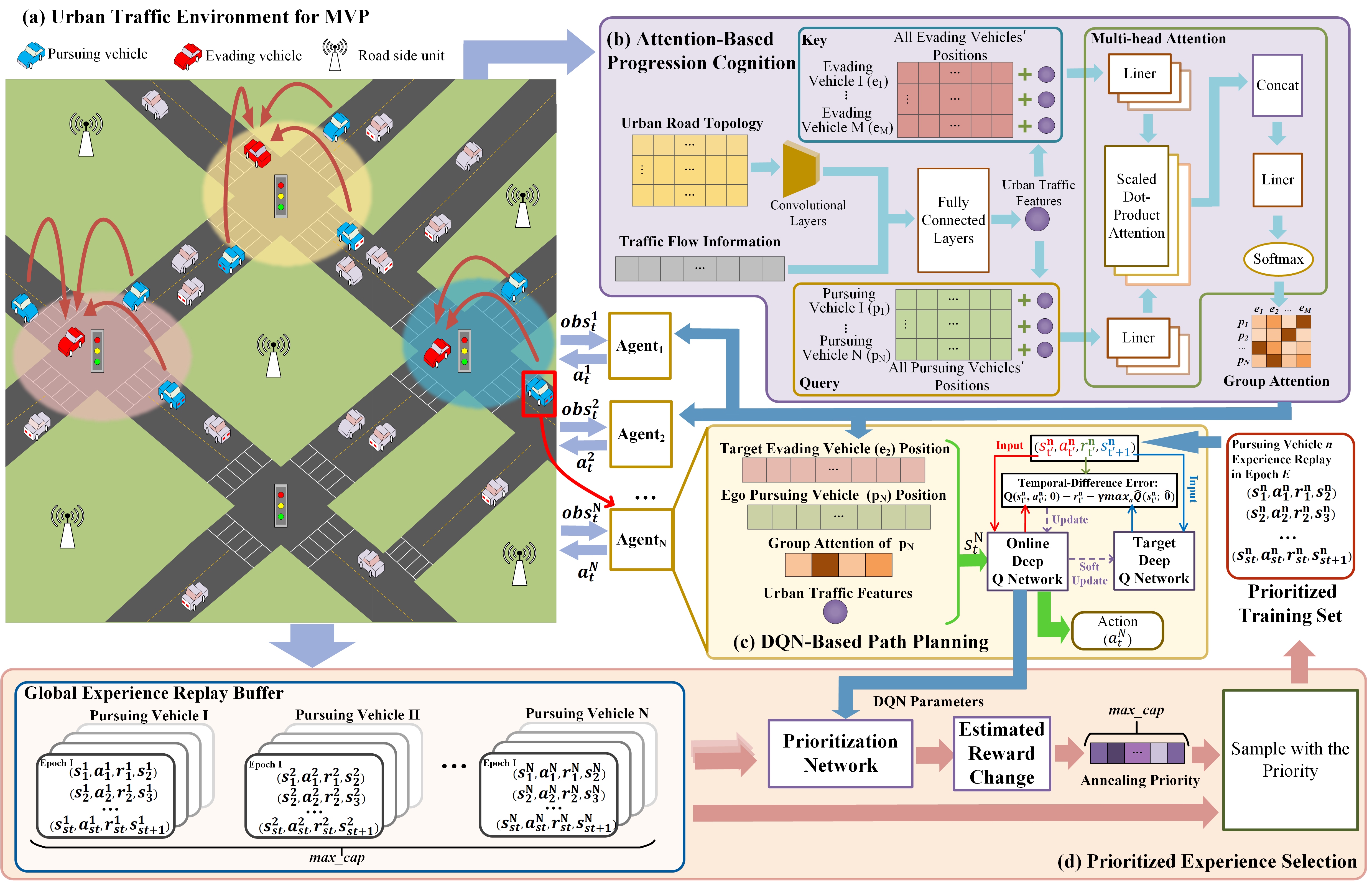}}
\caption{Architecture of PEPCRL-MVP. Urban traffic environment for MVP (\textbf{a}) provides complex pursuit-evasion scenes and interactive environment for MARL. Attention-based progression cognition module (\textbf{b}) provides accurate urban traffic information with critical features and group attention. The critical features and group attention are used to improve DQN-based path planning (\textbf{c}). Prioritization network and prioritized experience selection (\textbf{d}) are used to improve diversity and personalization of MARL. 
}
\label{fig_1}
\end{figure*}

Cooperative multi-agent reinforcement learning (MARL) has been widely studied for multiple agents collaboration and connected-automated vehicles (CAVs), and could be applied to MVP applications \cite{ad1,ad2,ad8}. Many MARL-based cooperative control schemes for CAVs have been proposed. Guan et al. presented a centralized coordination framework \cite{ad3} for autonomous vehicles at intersections without traffic signals, which significantly improved the road efficiency. \cite{ad4} and \cite{ad5} studied distributed cooperation methods to realize the conflict-free control of CAVs. \cite{ad6} and \cite{ad7} implemented multi-agent systems based hierarchical controller to improve vertical and horizontal cooperation among the automated vehicles. It is noted that all the above MARL algorithms for CAVs were designed to improve driving safety. However, the MVP tasks have additional mission critical requirements and require strong collaboration and adaptation to dynamic environments, which present significant new challenges to the design of MARL algorithms.


\IEEEpubidadjcol
In the literature, a few game theory-based MVP methods have been proposed \cite{mvp1}. Huang et al. presented a decentralized control scheme \cite{mvp2} based on the Voronoi partition of the game domain. \cite{mvp3} and \cite{mvp4} introduced curriculum RL to train pursuers to approach the evader. In order to improve the pursuing efficiency, \cite{mvp7} weighted different evaders to encourage the pursuers to capture the close evader. In addition, \cite{mvp5} constructed cooperative multi-agent schemes with target prediction network. Yang et al. design a graded-Q RL framework \cite{mvp9} to enhance the coordination capacity of pursuing vehicles. \cite{mvp6} and \cite{mvp8} adopted MARL to accomplish collaborative pursuit tasks in simplified traffic scenes with structured grid-pattern roads. However, these above methods did not consider the dynamic urban pursuit-evasion environment and the fixed allocation of pursuit tasks greatly affects the efficiency of the pursuit.

In the existing MARL algorithms for MVP and AD, deep neural network parameters are shared among agents via centralized training with decentralized execution (CTDE), which significantly improves learning efficiency and experience utilization \cite{marl1,marl2,marl3}. Many CTDE-based deep MARL methods achieve state-of-the-art performance on some tasks, such as group matching game and path finding \cite{marl0,marl4,marl5,marl6}. Although CTDE can accelerate training \cite{marl7}, it has poor performance in complex and difficult tasks, such as Google Research Football \cite{marl8} and MVP. These complex tasks typically require substantial exploration, diversified strategies, and efficient collaboration among agents \cite{marl9}. But agents tend to behave similarly because of parameter sharing, limiting efficient exploration and collaboration of MARL agents. 

According to the above analyses, it can be observed that low adaptation to dynamic traffic environments and homogeneous agents severely limit the collaborative pursuing performance. To address these problems, this paper proposes a progression cognition reinforcement learning with prioritized experience (PEPCRL-MVP) for MVP in urban traffic scenes. A framework of the PEPCRL-MVP is shown in Fig. \ref{fig_1}. There are two distinct modules in the new PEPCRL-MVP architecture. The first is a proposed prioritization network, which is used to select prioritized training set for each agent in MARL to adjust its deep neural network parameters. Optimizing agents with the personalized training set enables each agent to distinguish itself from others, thereby encouraging efficient collaboration. In addition, an attention-based progression cognition module is designed to adaptively group multiple pursuing vehicles considering dynamic traffic awareness. With the above designs, the PEPCRL-MVP can address the problems of low adaptation and homogeneous agents in the existing MVP approaches and is expected to greatly improve pursuing performance. 

The contributions of this paper can be summarized as follows.
\begin{itemize}
\item This paper proposes a multi-agent reinforcement learning approach with prioritized experience for collaborative multi-vehicle pursuit. A prioritization network is introduced to diversify the optimization and strategies of MARL, encouraging more efficient collaboration and experience exploration.
\item An attention-based progression cognition module is proposed to divide pursuing vehicles into improvisational groups according to dynamic urban traffic situations. Critical features are extracted from the sensor data to support the grouping process, which supports vehicles more effectively focusing on one target evading vehicle and greatly improves pursuing performance. 
\item This paper applies PEPCRL-MVP to the simulated urban large-scale roads with 46 junctions and sets different pursuing difficulty levels with variable numbers of pursuing vehicles and evading vehicles. In the three tested difficulty levels, PEPCRL-MVP improves pursuing efficiency by 3.95$\%$ on average compared with TD3-DMAP, and improves pursuing success rate by 34.78$\%$ on average compared with MADDPG. Codes are open sourced in \emph{https://github.com/BUPT-ANTlab/PEPCRL-MVP}.
\end{itemize}

The rest of this paper is organized as follows. Section \ref{formulation} describes multi-vehicle pursuit in an urban pursuit-evasion scene and models MVP problem based on partially-observable stochastic game (POSG). Section \ref{PEBMARL} presents MARL with prioritized experience and its training process. Section \ref{PathPlanning} presents the reinforcement learning-based path planning algorithm with progression cognition. Section \ref{results} gives the performance of the proposed method. Section \ref{conclu} draws conclusions.

\section{Multi-Vehicle Pursuit in Dynamic Urban Traffic} \label{formulation}
This section firstly illustrates the urban complex pursuit-evasion environment and the constraints, bridging the ‘sim-to-real’ gap. Section \ref{POSG} formulates the MVP problem based on POSG and introduces MARL-based solution to POSGs.  

\subsection{MVP in Large-Scale Urban Traffic}
This paper focuses on the problem of multi-vehicle pursuit under the complex urban traffic. We consider a closed large-scale urban traffic scene with multi-intersection road structure \cite{SUMO}. The considered scene basically retains the settings of urban traffic, such as traffic light and speed limits. Without loss of generality, we assume there are $N$ pursuing vehicles, $M$ evading vehicles ($N>M$), $B$ background vehicles, and $L$ lanes. The background vehicles and evading vehicles follow the randomly selected routes. For the MVP task, an evading vehicle is deemed captured if any pursuing vehicle is less than a pre-configured distance $d_{min}$ from its target evading vehicles. If all evading vehicles are captured with a given $st$ time steps, the pursuit task is successfully \emph{Done}. 

In this paper, we have a few constraints for the MVP task.
\begin{itemize}
\item  All vehicles in the scene (pursuing, evading, and background vehicles) adhere to traffic rules, such as obeying traffic lights, and drive without collisions.
\item All pursuing vehicles and evading vehicles are initialized at different diagonal points in the map with $0 \text{ }m/s$.
\item The maximum speed $\mathop v\nolimits_{\max}$, maximum acceleration $\mathop {ac}\nolimits_{\max}$ and maximum deceleration $\mathop {de}\nolimits_{\max}$ of all pursuing vehicles and evading vehicles are set to be the same.
\end{itemize}

\subsection{POSG-Based MVP Problem Formulation} \label{POSG}
In MVP, the decision-making process of a finite set of agents ${\cal I}$ deployed in pursuing vehicles with partial observability can be formalized as POSG, which can be defined as a tuple ${\mathop {\cal M}\nolimits_G : = ({\cal I},{\cal S},[\mathop {\cal A}\nolimits_n ],[\mathop {\cal O}\nolimits_n ],Tr,[\mathop R\nolimits_n ])}$ for ${n = 1,...,N}$. In time step $t$, the pursuing vehicle $n$ receives a local observation ${\mathop o\nolimits_t^n :{\cal S} \to \mathop {\cal O}\nolimits_n}$ that is correlated with the underlying state of the environment ${\mathop s\nolimits_t  \in {\cal S}}$. ${\mathop o\nolimits_t^n}$ is further processed to ${\mathop s\nolimits_t^n}$ as the state of the pursuing vehicle $n$, that takes an action ${a_t^n \in \mathop {\cal A}\nolimits_n}$ according to ${\mathop s\nolimits_t^n}$. Consequently, the environment evolves to a new state ${\mathop s\nolimits_{t+1}}$ with the transition probability ${Tr = P(\mathop s\nolimits_{t + 1} |\mathop s\nolimits_t ,\mathop a\nolimits_t ):{\cal S} \times \mathop {\cal A}\nolimits_1  \times ... \times \mathop {\cal A}\nolimits_N  \to {\cal S}}$ and then the agent receives a decentralized reward ${r_t^n:\mathcal{S} \times \mathop \mathcal{A}\nolimits_n  \to \mathbb{R}}$. The probability distribution of actions at a given state is determined by the stochastic policy ${\mathop \pi \nolimits_n}$. The goal of an optimal policy ${\mathop \pi \nolimits_n^*}$ is to generate a distribution that maximizes the discounted sum of future rewards over an infinite time horizon, which can be expressed as

\begin{equation}
\pi _n^*: = \mathop {\arg \max }\limits_\pi  \mathbb{E}(\sum\limits_{t = 0}^\infty  {\mathop \gamma \nolimits^t } R({\rm{ }}s_t^n,\pi (s_t^n))),
\end{equation}
in which, ${\gamma  \in [0,1)}$ is the discount factor, indicating the impact of future earnings on current expectation value. The optimal policy maximizes the state-action value function, i.e., ${\pi _n^*(s_t^n) = \mathop {\arg \max }\limits_a \mathop Q\nolimits_n^\pi  (s_t^n,a)}$. 

According to Bellman optimality equation, the optimal state-action value function can then be derived as
\begin{equation}
{\mathop Q\nolimits_n^\pi  (s_t^n,a_t^n) = \mathop \mathbb{E} \nolimits_{\mathop s\nolimits_{t + 1} \sim P(.|\mathop s\nolimits_t ,\mathop a\nolimits_t )} (r_t^n + \mathop {\gamma \rm{argmax} }\limits_a \mathop Q\nolimits_n^\pi  (s_{t + 1}^n,a))}.
\end{equation}

As an emerging AI algorithm, MARL enables agents in POSGs to make optimal strategies without exact state transition probability $Tr$ and reward function $R$. It provides an excellent solution to MVP with the dynamic and complex environment. The agent $n$ in MARL updates the Q value function according to temporal difference error $\delta$ by off-policy learning, 
 \begin{equation}
{\delta  = r_t^n + \gamma \mathop {\arg \max }\limits_{{a_{t + 1}}} Q_n^\pi (s_{t + 1}^n,a_{t + 1}^n) - Q_n^\pi (s_t^n,a_t^n)},
 \end{equation}
 
\begin{equation}
\label{UD-Q}
{Q_n^\pi (s_t^n,a_t^n) \leftarrow Q_n^\pi (s_t^n,a_t^n) + \alpha \delta},
\end{equation}
where $\alpha$ is the learning rate. For pursuing vehicle $n$, the function $Q_n^\pi (s_t^n,a_t^n)$ calculates the expectation values of turning left, turning right and going straight according to the current state $s_t^n$ to assist the vehicle to select the optimal route to pursue the evading vehicle.

\section{MARL with Prioritized Experience} \label{PEBMARL}
To introduce diversity among collaborative agents, we design a prioritized experience boosting MARL equipped with a prioritization network. Subsection \ref{framework} describes the overall framework and Subsection \ref{PN} presents the prioritization network in detail. Finally, the training process is introduced.

\subsection{Prioritized Experience Boosting MARL Framework}\label{framework}
\begin{figure*}[]
\centering
\centerline{\includegraphics[width=0.98\textwidth,height=8.3cm]{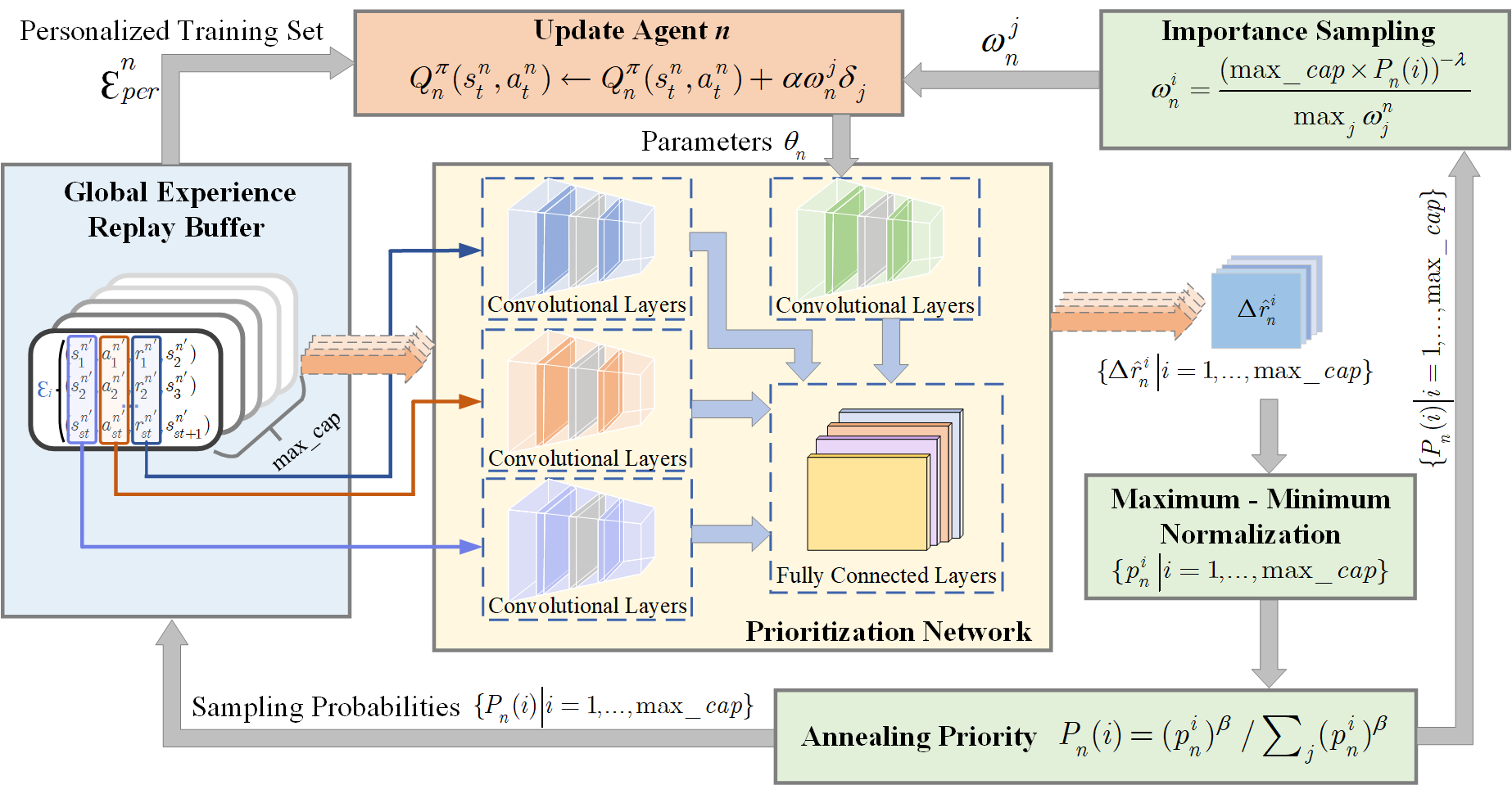}}
\caption{Architecture of prioritized experience boosting MARL. 
}

\label{fig_2}
\end{figure*}

Emerging MARL algorithms, such as MADDPG\cite{marl1} and QMIX\cite{qmix}, adopt centralized training with randomly sampling experience to improve the utilization of experience and deploy the same trained model to all the agents. 
However, homogeneous learning policy hinders collaboration among agents. Moreover, randomly sampling experience also affects the training efficiency. Therefore, this paper proposes a prioritized experience boosting MARL framework, as shown in Fig. \ref{fig_2}, to introduce diversity among agents.
It employs a prioritization network $PN$ to select personalized training set $\mathop {\cal E}\nolimits_{per}^n  \in {\cal G}$ by a central server from the global experience replay buffer ${\cal G} = \{ \mathop {\cal E}\nolimits_1 ,\mathop {\cal E}\nolimits_2 ,...,\mathop {\cal E}\nolimits_{{\rm{max\_cap}}} \}$ for each agent. ${max\_cap}$ represents the maximum capacity of the global buffer, and ${\cal E} $ is the replay experience collected by agent $n'$ in one epoch,

\begin{equation}
    {\cal E} = \{ (s_1^{n'},a_1^{n'},r_1^{n'},s_2^{n'}),...,(s_{st}^{n'},a_{st}^{n'},r_{st}^{n'},s_{st+1}^{n'})\}.
\end{equation}

In the prioritized experience boosting MARL framework, all agents upload exploration experience to the central server. Every agent samples prioritized experience for training and updates parameters in the global experience buffer via the prioritization network. The prioritization network is trained to model the relationship between the training set features and the reward change $\Delta r$ after updating parameters, thus assisting the agents in selecting appropriate experience for efficient training and improving decision performance.

The prioritized experience boosting MARL essentially optimizes the gradient descent and parameter update process of RL-based agents. It differs from the conventional training way of simply replaying experiences at the same frequency, regardless of their importance. By prioritizing experiences based on their significance, the framework enables agents to train more efficiently with optimal replay transitions and fosters better collaboration among the agents.


\subsection{Prioritization Network and Annealing Priority}\label{PN}
The prioritization network is responsible for assessing the importance of experience replay transition $\mathop {\cal E}\nolimits_i$ for all agents. 
For a given agent $n$ with parameters ${\mathop \theta \nolimits_n}$, the prioritization network determines the priority and the sampling probability $P_n(i)$ of experience replay transition $\mathop {\cal E}\nolimits_i$. 
Thus, the prioritization network is designed to estimate the performance gain of the agent $n$ after training with $\mathop {\cal E}\nolimits_i$. And, as for RL, the rewards directly reflect the performance of an agent. Therefore, we use the prioritization network to fit the reward change $\mathop {\Delta r}\nolimits_n^i $ after training the agent $n$ with experience replay transition $\mathop {\cal E}\nolimits_i$,

\begin{equation}
    \hat{\mathop {\Delta r}\nolimits_n^i}  = PN(\mathop {\cal E}\nolimits_i ,\mathop \theta \nolimits_n ;\vartheta ),
\end{equation}
where $\vartheta$ is the parameters of the prioritization network $PN$. 
Meanwhile, in order to improve the stability of the algorithm, we use the average reward over $k$ historical epochs as the base reward to calculate the reward change $\mathop {\Delta r}\nolimits_n^i$. We adopt gradient back-propagation to update $\vartheta$. The loss of $PN$ is defined as,
 \begin{equation}
 \label{equ_7}
     J(\vartheta ) = \frac{1}{N}\sum\limits_n {{{[PN(\mathop {\cal E}\nolimits_i ,\mathop \theta \nolimits_n ;\vartheta ) - \mathop {\Delta r}\nolimits_n^i ]}^2}}. 
 \end{equation}

The prioritization network output $ \hat{\mathop {\Delta r}\nolimits_n^i}$ is used for stochastic sampling. 
The prioritization network computes the gain of each replay transition in the global experience replay buffer according to the current parameters of agent $n$, which is defined as $\{ \hat {\Delta r_n^i}\left| {i = 1,...,} \right. max\_cap\}$. 
And then the maximum-minimum normalization is performed on the sequence,

\begin{equation}
\label{equ_8}
    \mathop q\nolimits_n^i  = \frac{{\hat {\Delta r_n^i} - \mathop {\min }\limits_j (\hat {\Delta r_n^j})}}{{\mathop {\max }\limits_j (\hat {\Delta r_n^j}) - \mathop {\min }\limits_j (\hat {\Delta r_n^j})}} + \zeta, 
\end{equation}
where $\zeta$ is a small positive constant that prevents the sampling probability becoming zero. However, reward prioritization sampling may focus on a small subset of the experience that makes the agent prone to over-fitting. Therefore, the annealing priority is proposed to calculate sampling probabilities,
\begin{equation}
\label{equ_9}
    {\mathop P\nolimits_n (i) = \frac{{\mathop {(\mathop q\nolimits_n^i )}\nolimits^\beta  }}{{\sum\limits_j {\mathop {(\mathop q\nolimits_n^j )}\nolimits^\beta  } }}},
\end{equation}
where the exponent $\beta  \in (0,1]$ determines how much prioritization is introduced. $\beta=0$ corresponds to the uniform sampling. In the early stage of training, uniform sampling is expected to facilitate agents learning and the prioritization network convergence. As training proceeds, the $PN$ can gradually  and reliably compute the value of an experience replay transition and guide the agents gradient decreasing. In practice, we linearly anneal $\beta$ from $\beta_0$ to 1 to ensure  stable MARL update and continuous performance improvement.

\subsection{Training Process of MARL with Prioritized Experience}
Typical reinforcement learning utilizes random experience replay to estimate the distribution of policy and states via Monte Carlo. However, prioritized replay inevitably changes the distribution and introduces bias \cite{PER}. And the optimal solution that the estimates converge to is influenced by the bias. 
Therefore, this paper adopts Importance Sampling (IS) to abate the impact of the bias,
\begin{equation}
    {\mathop \omega \nolimits_n^i  = \frac{{\mathop {({\rm{max\_cap}} \times \mathop P\nolimits_n (i))}\nolimits^{ - \lambda } }}{{\mathop {\max }\limits_j \mathop \omega \nolimits_n^j}}},
\end{equation}
that fully compensates for the non-uniform probabilities $P_n(i)$ if $\lambda=1$. 
And the weights is normalized by ${1/{\mathop {\max }\limits_j \omega _n^j}}$ for stability. 
It is worth noting that the choice of hyperparameters $\lambda$ interacts with $\beta$ in annealing priority. 
Increasing them both simultaneously encourages more aggressive priority sampling. Considering IS, Eq. \eqref{UD-Q} can be reformulated as, 
 \begin{equation}
 \label{IS}
     {Q_n^\pi (s_t^n,a_t^n) \leftarrow Q_n^\pi (s_t^n,a_t^n) + \alpha \mathop \omega \nolimits_n^i \mathop \delta }.
 \end{equation}

Introducing IS into the training process has another benefit. IS can reduce the step size of non-linear function approximation, e.g. deep neural networks. In prioritized experience boosting MARL, experience replay transitions favored by the prioritization network may be revisited many times, and the IS correction reduces the gradient magnitude to ensure that the agents converge to the globally optimal policy.

\begin{algorithm}[h]\label{alg_1}
            \caption{Training Process of Prioritized Experience Boosting MARL}
            
            \KwIn{$N$ agents, prioritization network $PN$, average reward of each agent in $k$ historical test epochs $[\mathop {\bar r}\nolimits_1 ,\mathop {\bar r}\nolimits_2 ,...,\mathop {\bar r}\nolimits_N ]$, and global experience replay buffer ${\cal G}$}
            \LinesNumbered

            \For{n=1:N}{
            \For{i=1:$max\_cap$}{
             Get the priority of ${\cal E}_i$ in ${\cal G}$ by $PN$ considering parameters $\theta_n$ of agent $n$\;
            }
            Calculate the annealing priority $P_n$ for each replay transition via Eq. \eqref{equ_8} and Eq. \eqref{equ_9}\;
            Sample personalized training set $\mathop {\cal E}\nolimits_{per}^n$ for agent $n$ according to $P_n$\;
            Update parameters $\theta_n$ of agent $n$ via Eq. \eqref{IS}\;
            }
            Test the updated MARL and get distributed rewards $[\mathop {r}\nolimits_1 ,\mathop {r}\nolimits_2 ,...,\mathop {r}\nolimits_N ]$ \;
            Calculate the change of rewards $[\mathop {\Delta r}\nolimits_1 ,\mathop {\Delta r}\nolimits_2 ,...,\mathop {\Delta r}\nolimits_N ]$ \;
            Calculate the loss of $PN$ via Eq. \eqref{equ_7} and update $PN$\;
    \end{algorithm}
    
Distributed training is used in the proposed prioritized experience boosting MARL. The overall training process is shown in Algorithm \ref{alg_1}. 
For every RL-based agent, the prioritization network firstly evaluates the priority of each experience transition according to the agent's parameters. And the annealing priority is obtained to help select personalized training set $\mathop {\cal E}\nolimits_{per}^n$. Then the agent's parameters are updated with $\mathop {\cal E}\nolimits_{per}^n$ via Eq. \eqref{IS}. After the parameters of all the agents have been updated, the prioritized experience boosting MARL is tested, and the gain of each agent's reward is calculated. Finally, the prioritization network is trained via Eq. \eqref{equ_7}. With this training process, the prioritization network can accurately compute the value of experience replay transitions for each agent. Moreover, by training with personalized and optimal experience, diverse multiple agents can largely improve collaboration and convergence efficiency.

\section{Progression Cognition DQN-Based Cooperative Path Planning}\label{PathPlanning}

This section will introduce a progression cognition DQN-based cooperative path planning for pursuing vehicles. 
Firstly, the attention-based progression cognition module is presented in Section \ref{ProactiveGroupCognition}. 
DQN-based path planning, as the core decision making algorithm, is then described in Section \ref{MVPPP}. 
Finally, Section \ref{process} introduces the decision-making and training process of the proposed PEPCRL-MVP.

\subsection{Attention-Based Progression Cognition Module}\label{ProactiveGroupCognition}

In complex urban traffic environments pursuing vehicles need real-time and accurate sensing of driving environments and the status of evading vehicles. We propose attention-based progression cognition module to extract critical traffic features and assist pursuing vehicles to select suitable evading vehicle as the target. 
It helps each pursuing vehicle focus on only one evading vehicle and work with other pursuing vehicles in a group to improve pursuit performance. 
Moreover, the allocation of pursuing tasks with progression cognition enhance collaboration among pursuing vehicles.

The locations of the pursuing and evading vehicles are very important for collaboration and decision making of the pursuing vehicles. 
In this paper, the location of vehicle $i$ is denoted by ${loc}^i_t = \left\{ {{C_l},pos^{i,l}_t} \right\}$. 
${C_l}$ denotes the binary code of lane $l$ where vehicle $i$ is located, 
and $pos^{i, l}_{t}$ denotes the distance between vehicle $n$ and the starting of lane $l$ at time $t$. 
The positions of pursuing and evading vehicles can be represented respectively as $\mathop {{\cal L}{\cal O}{\cal C}\_{\cal P}}\nolimits_t  = \{ loc_t^1,loc_t^2,...,loc_t^N\} $ and $\mathop {{\cal L}{\cal O}{\cal C}\_{\cal E}}\nolimits_t  = \{ loc_t^1,loc_t^2,...,loc_t^M\} $. 
Moreover, the adjacency matrix $RT$ is used to represent the topology of roads. 
Assume that there are $L$ lanes in the pursuit-evasion environment. 
The size of matrix $RT$ is $L \times L$. An element $e_{i,j}$ in row $i$ and column $j$ of $RT$ indicates whether the vehicles can drive directly from lane $i$ to lane $j$.


To choose an optimal pursuit route, the information of the number of background vehicles in each lane  is also utilized by the progression cognition module. 
The number of background vehicles in each lane forms a vector of size $1 \times L$, defined as $BV_t$. 
We use convolutional neural networks (CNNs) in the module to extract key traffic features from $RT$ and $BV_t$. 
Specifically, $RT$ is fed into the convolutional layers. Then its output combined with $BV_t$ is input to the fully connected layers, and finally the urban traffic feature $F$ is obtained.

As the core of progression cognition module, multi-head attention is used to help pursuing vehicles focus on evading vehicles. It simultaneously takes into account urban traffic features $F$. All pursuing vehicles share their locations. And $F$ is attached to the vehicle position vectors ${loc}^i_t$, which is a word embedding. Therefore, the query $Q$ of multi-head attention is represented as $[[loc_t^1,F],...,[loc_t^N,F]]$ and the key $K$ is represented as $[[loc_t^1,F],...,[loc_t^M,F]]$. Group attention weights $W_g$ can be derived as

\begin{equation}
    \mathop W\nolimits_g  = \mathop \psi \nolimits^{liner} (Concact(\mathop W\nolimits_1 ,\mathop W\nolimits_2 ,...,\mathop W\nolimits_h )),
\end{equation}
\emph{in which,}
\begin{equation}
    \mathop W\nolimits_i  = Softmax(\frac{{\mathop {\rm{Q}}\nolimits_i {\rm{ }}\mathop K\nolimits_i^T }}{{\sqrt {\mathop d\nolimits_K } }}),
\end{equation}

Here, $h$ is the number of heads and ${\mathop d\nolimits_K }$ is the dimension of $K$'s features. The size of group attention weights $W_g$ finally obtained is $N \times M$. The $n$th row of $W_g$ represents the attention weight of pursuing vehicle $n$ on all evading vehicles, where a larger value means more attention. Every pursuing vehicle selects an evading vehicle with the maximum attention weight as its target vehicle. Due to the high dynamic of the urban traffic, the target evading vehicle selected by pursuing vehicle $n$ may vary from time steps. Therefore, the progression cognition divides the pursuing vehicles adaptively to collaborative groups according to the traffic situations.  


\subsection{Multi-Vehicle Pursuit Path Planning}\label{MVPPP}
\begin{figure}[]
\centering
\centerline{\includegraphics[width=0.88\columnwidth]{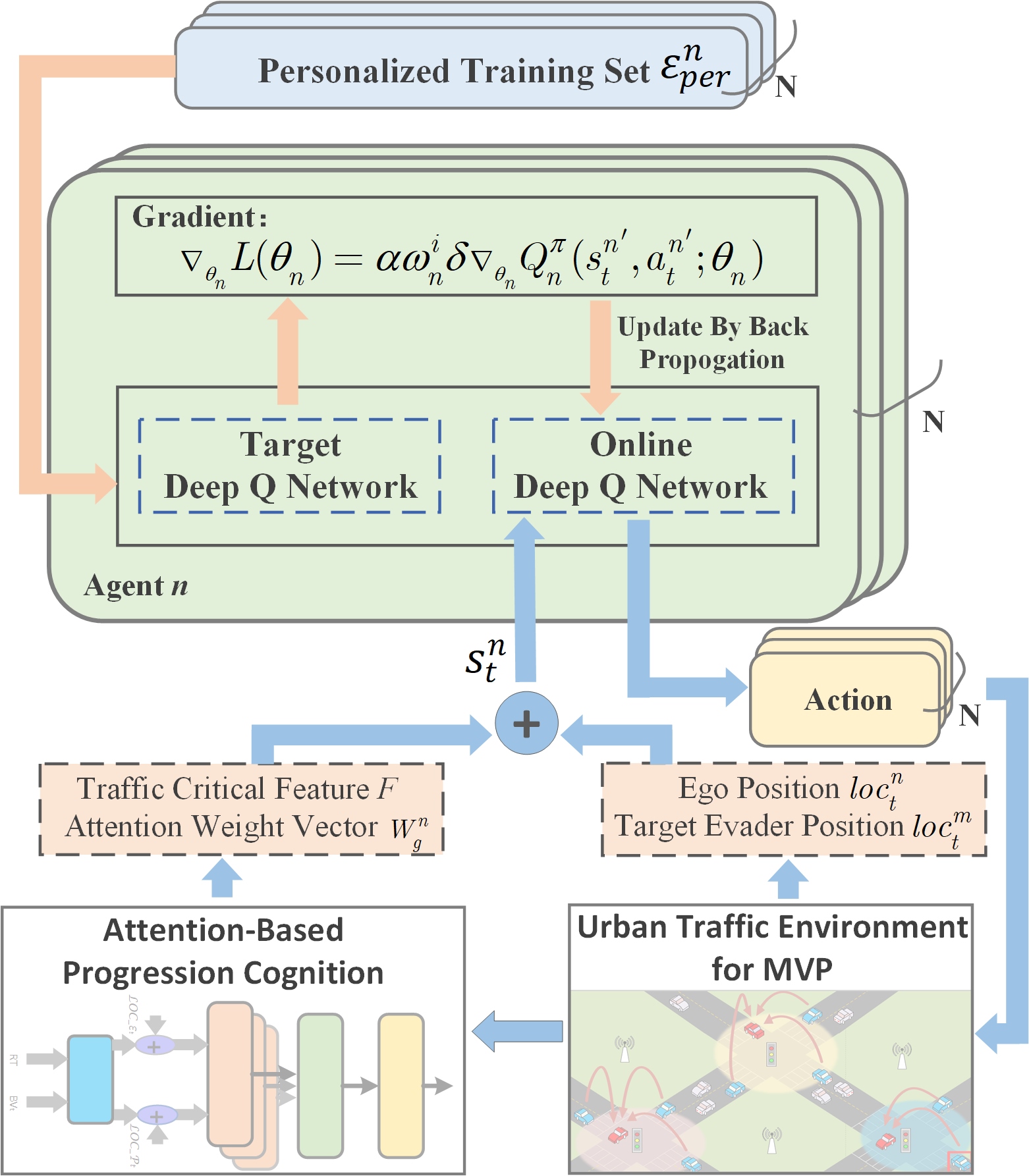}}
\caption{Architecture of DQN-based multi-vehicle pursuit path planning. 
}
\label{fig_3}
\end{figure}

Deep Q-Network (DQN) is a popular reinforcement learning algorithm and has been applied for decision-making in various scenarios. In DQN, artificial neural networks (ANNs) are used to approximate $Q_n^\pi (s_t^n,a_t^n)$. 
DQN adopts a dual network framework, consisting of ${online}$ network and ${target}$ network, which have the same structure. 
The two ANNs are parameterized by ${\theta _n}$ and ${\theta _n}'$. 
In MVP path planning, DQN is utilized to evaluate the value $Q$ of the pursuing vehicle's each action according to its real-time state $s_t^n$, 
denoted as $Q_n^\pi (s_t^n\left| {{\theta _n}} \right.)$. $Q_n^\pi (s_t^n\left| {{\theta _n}} \right.)$ is used as the policy for agent $n$ to select the appropriate action. The architecture of multi-vehicle pursuit path planning algorithms is shown in Fig. \ref{fig_3}.

In this paper, the action space of each pursuing vehicle includes three actions, turning left, turning right and going straight. 
For agent $n$, the state $s_t^n$ consists of four parts, including its own position ${loc}^n_t$, the position of the target evading vehicle being focused on ${loc}^m_t$, the traffic critical feature $F$ and its attention weight vector $W_g^n$. 
Here, $W_g^n$ represents the attention weight of agent $n$ on all evading vehicles obtained via attention-based progression cognition module, which is the $n$th row of $W_g$. 

To motivate the capture of pursuing vehicles and incentivize efficient training, a carefully designed reward $r_t^n$ consists of \emph{sparse reward} and \emph{dense reward}. 
\textbf{\emph {Sparse Reward}}: Only when a collaborative group successfully captures its target, all members in the group obtain a positive reward $V$.
\textbf{\emph {Dense Reward}}: Sparsity of reward hinders the exploration of optimal policy by agents. Therefore, a \emph{dense reward} is set for each pursuing vehicle to address this problem. The dense reward contains the following two components, 1) The pursuing vehicles which do not capture the target vehicle are given a negative reward of $c$ at each time step; 2) A distance-sensitive reward is set to improve the pursuing efficiency. 
When a pursuing vehicle reduces the distance from its current target compared to that at the last time step, it will obtain a positive reward, and conversely, it will be punished with a negative reward. 

Therefore, the formulation of $r_t^n$ is expressed as
\begin{equation}
r_t^n = \left\{ {\begin{array}{*{20}{l}}
V&{\text{if successful pursuit}}\\
{c \times t + \sigma \left( {d_t^{n,m} - d_{t - 1}^{n,m}} \right)}&{{\rm{else }}}
\end{array}} \right. ,
\end{equation}
where $\sigma$ is a negative reward factor, and $d^{n, m}_{t}$ denotes the distance of the pursuing vehicle $n$ from its target evading vehicle $m$ at time step $t$. 

Adopting prioritized experience boosting MARL, we can compute the following gradient by differentiating the loss function with respect to the weights,

\begin{equation}
\label{eq_16}
    \mathop \nabla \nolimits_{{\theta _n}} L({\theta _n}) = \alpha {\rm{ }}\mathop \omega \nolimits_n^i \delta \mathop \nabla \nolimits_{{\theta _n}} Q_n^\pi (s_t^{n'},a_t^{n'};{\theta _n}),
\end{equation}
\emph{in which,}
\begin{equation}
    \delta  = r_t^{n'} + \gamma \mathop {\max }\limits_{a_{t + 1}^{n'}} {\rm{ }}Q_n^{\pi '}(s_{t + 1}^{n'},a_{t + 1}^{n'};{\theta _n}') - {\rm{ }}Q_n^\pi (s_t^{n'},a_t^{n'};{\theta _n}),
\end{equation}
\begin{equation}
    (s_t^{n'},a_t^{n'},r_t^{n'},s_{t + 1}^{n'}) \in \mathop {\cal E}\nolimits_{per}^n ,
\end{equation}

${\theta _n}$ is updated via stochastic gradient descent and Eq. \eqref{eq_16}. And ${target}$ network performs soft update at each training step,
\begin{equation}
\label{eq_20}
    {\theta _n}' \leftarrow \tau {\theta _n} + (1 - \tau ){\theta _n}'.
\end{equation}

\subsection{PEPCRL-MVP Decision-Making and Training Process}\label{process}

The decision-making and training process of the proposed PEPCRL-MVP is shown in Algorithm \ref{alg_2}. $N$ DQN-based path planning agents and the prioritization network are initialized firstly. At the beginning of each epoch, $N$ agents are trained distributedly with personalized and prioritized experience via Algorithm \ref{alg_1}, if the number of transitions in the global experience replay buffer ${\cal G}$ reaches $max\_cap$. Then the pursuit-evasion environment is initialized to test PEPCRL-MVP. In each time step of pursuit, attention-based progression cognition processing is invoked to get $F$ and $W_g$. Then each agent uses the partial observation to obtain optimal path planning and the strategy of all agents is performed. At the end of the epoch, experience of all pursuing vehicles is stored to ${\cal G}$. Finally, change of rewards is calculated and used to update the prioritization network $PN$ via Eq. \eqref{equ_7}.

\begin{algorithm}[h]\label{alg_2}
            \caption{PEPCRL-MVP Decision-making and Online Training Algorithm}
            
            \LinesNumbered

            Initialize $N$ DQN-based path planning agents and the prioritization network $PN$\;
            Initialize global experience replay buffer ${\cal G}$ \;
           \For{e=1:max\_epoch}{
           \If{$len({\cal G}) = max\_cap$}{
            Train $N$ agents via Algorithm \ref{alg_1}, Eq. \eqref{eq_16} \;
           }
           Initialize an urban pursuit-evasion environment and obtain $S_1=\{ {\mathop {{\cal L}{\cal O}{\cal C}\_{\cal P}}\nolimits_1}, {\mathop {{\cal L}{\cal O}{\cal C}\_{\cal E}}\nolimits_1}, BV_1 \} $\;
           Get $F$ and $W_g$ by progression cognition\;
           
           \For{t=1:st}{ 
           \For{n=1:N}{
           $s_t^n \leftarrow [loc_t^n,loc_t^m,F,W_g^n]$\;
           Obtain $a_t^{n}$ by $Q_n^\pi (s_t^n\left| {{\theta _n}} \right.)$\;
           }
           Perform the strategy $a_t$ and observe $S_{t+1}$\;
           Get $F$ and $W_g$\ by progression cognition\;
           Append $(s_t^{n},a_t^{n},r_t^{n},s_{t + 1}^{n})$ to $\mathop {\cal E}\nolimits_n$, $n=1,...,N$\;
           $ S_{t}  \leftarrow S_{t+1}$ \;
           \If{$Done$}{
           break\;
           }
           }
           Store $\mathop {\cal E}\nolimits_n$ to ${\cal G}$, $n=1,...,N$\;
           Get per step average rewards $[\mathop {r}\nolimits_1 ,\mathop {r}\nolimits_2 ,...,\mathop {r}\nolimits_N ]$\;
           \If{$len({\cal G}) = max\_cap$}{
           Calculate change of rewards and update $PN$\;
           }
           Update mean rewards of $k$ epochs $[\mathop {\bar r}\nolimits_1 ,\mathop {\bar r}\nolimits_2 ,...,\mathop {\bar r}\nolimits_N ]$\;

           }
    
\end{algorithm}

\section{Experiments and Results}\label{results}
\subsection{The Simulator and Settings}
To comprehensively evaluate the proposed PEPCRL-MVP, we build three urban traffic road scenes with bidirectional two lanes based on SUMO \cite{SUMO}, including $3 \times 3$ and $4 \times 5$ grid-pattern urban roads, and real map-based urban roads. The real map-based urban roads simulate those in an area inside the second ring road of Beijing, bridging the ‘sim-to-real’ gap. During the simulation process, the number of background vehicles remains constant, and the background vehicles follow randomly selected routes. Moreover, to evaluate the robustness of PEPCRL-MVP, we design three different difficulty levels of MVP tasks with variable numbers of pursuing vehicles $N$ and evading vehicles $M$, respectively, 6 pursuing vehicles chasing 3 evading vehicles (denoted by P6-E3), 7 pursuing vehicles chasing 4 evading vehicles (denoted by P7-E4) and 8 pursuing vehicles chasing 5 evading vehicles (denoted by P8-E5). All evading vehicles randomly select escape routes. The simulation parameters are shown in TABLE \ref{SimuPar} and the  PEPCRL-MVP parameters are shown in TABLE \ref{SetPar}.


\begin{table}[h]
\centering
\caption{Simulator Settings}
\begin{tabular}{cc|c}
\hline
\multicolumn{2}{c|}{Parameters}                                                                                         & Values \\ \hline
\multicolumn{2}{c|}{maximum time steps $st$}                                                                               & 800    \\
\multicolumn{2}{c|}{capture distance $d_{min}$}                                                                               & 5 $m$ \\
\multicolumn{2}{c|}{maximum speed $\mathop v\nolimits_{\max}$}                                                                               & 20 $m/s$    \\
\multicolumn{2}{c|}{maximum acceleration $\mathop {ac}\nolimits_{\max}$}                                                                       & 0.5 $m/s^2$   \\
\multicolumn{2}{c|}{maximum deceleration $\mathop {de}\nolimits_{\max}$}                                                                       & 4.5  $m/s^2$  \\ \hline
\multirow{4}{*}{\begin{tabular}[c]{@{}c@{}}3×3\\ urban roads\end{tabular}} & number of lanes $L$             & 48     \\
                                                                                        & number of junctions           & 16     \\
                                                                                        & length of each lane           & 500 $m$    \\
                                                                                        & number of background vehicles & 240    \\ \hline
\multirow{4}{*}{\begin{tabular}[c]{@{}c@{}}4×5\\  urban roads\end{tabular}} & number of lanes $L$             & 98     \\
                                                                                        & number of junctions           & 30     \\
                                                                                        & length of each lane           & 400 $m$    \\
                                                                                        & number of background vehicles & 500    \\ \hline
\multirow{3}{*}{\begin{tabular}[c]{@{}c@{}}real map-based \\ urban roads\end{tabular}}  & number of lanes $L$             & 106    \\
                                                                                        & number of junctions           & 46     \\
                                                                                        & number of background vehicles & 300    \\ \hline
\end{tabular}
\label{SimuPar}
\end{table}

\begin{table}[h]
\centering
\caption{Parameter Settings}
\begin{tabular}{ll|ll}
\hline
Parameters    & Values & Parameters & Values \\ \hline
$\alpha$ &   $\mathop {10}\nolimits^{ - 4}$   & $V$          & 400    \\
$\gamma$         & 0.9   & $c$          & 0.02   \\
$\beta_0$        & 0.01 & $\sigma$       & 5      \\
$\lambda$        & 0.5    & $\tau$       & 0.001 \\ \hline
\end{tabular}
\label{SetPar}
\end{table}

\begin{table*}[]
\caption{Evaluation Results}
\begin{center}
\renewcommand{\arraystretch}{1.162}
\begin{tabular}{cccccccccccc}
\toprule
\begin{tabular}[c]{@{}c@{}}Road\\Structure\end{tabular}                & N-M                  & \begin{tabular}[c]{@{}c@{}}Evaluate\\ Metrics\end{tabular} & \begin{tabular}[c]{@{}c@{}}PEPCRL-\\ MVP\end{tabular} & A-MVP       & B-MVP                            & DQN     & DDPG    & MADDPG  & \begin{tabular}[c]{@{}c@{}}TD3-\\ DMAP\end{tabular} & PPO     & T$^{3}$OMVP  \\ \toprule
\multirow{15}{*}{3×3}      & \multirow{5}{*}{P6-E3} & \multicolumn{1}{c|}{AR}                                    & 3.164                                                 & 2.463   & \multicolumn{1}{c|}{2.558}   & 2.154   & 1.461   & 1.552   & 2.252                                               & 2.247   & 0.698   \\
                           &                      & \multicolumn{1}{c|}{SDR}                                   & 6.927                                                 & 7.656   & \multicolumn{1}{c|}{8.069}   & 7.179   & 6.549   & 6.831   & 6.919                                               & 7.831   & 5.701   \\
                           &                      & \multicolumn{1}{c|}{ATS}                                   & 642.87                                                & 684.77  & \multicolumn{1}{c|}{655.40}   & 677.35  & 700.75  & 693.26  & 676.90                                               & 687.30   & 717.61  \\
                           &                      & \multicolumn{1}{c|}{SDTS}                                  & 175.410                                                & 173.669 & \multicolumn{1}{c|}{189.742} & 179.388 & 166.467 & 171.972 & 176.389                                             & 181.518 & 136.882 \\
                           &                      & \multicolumn{1}{c|}{SR}                                    & 0.62                                                  & 0.48    & \multicolumn{1}{c|}{0.51}    & 0.47    & 0.36    & 0.42    & 0.48                                                & 0.49    & 0.37    \\ \cline{2-12} 
                           & \multirow{5}{*}{P7-E4} & \multicolumn{1}{c|}{AR}                                    & 2.183                                                 & 1.407   & \multicolumn{1}{c|}{1.915}   & 1.372   & 0.554   & 0.974   & 0.608                                               & 1.365   & 0.746   \\
                           &                      & \multicolumn{1}{c|}{SDR}                                   & 6.694                                                 & 6.482   & \multicolumn{1}{c|}{7.004}   & 6.152   & 5.889   & 5.740    & 5.617                                               & 6.082   & 5.711   \\
                           &                      & \multicolumn{1}{c|}{ATS}                                   & 684.50                                                 & 692.77  & \multicolumn{1}{c|}{688.13}  & 695.89  & 738.15  & 726.81  & 735.03                                             & 696.56  & 700.54  \\
                           &                      & \multicolumn{1}{c|}{SDTS}                                  & 150.082                                               & 149.366 & \multicolumn{1}{c|}{159.150}  & 153.201 & 119.352 & 132.714 & 120.750                                              & 152.391 & 145.638 \\
                           &                      & \multicolumn{1}{c|}{SR}                                    & 0.55                                                  & 0.46    & \multicolumn{1}{c|}{0.49}    & 0.45    & 0.35    & 0.37    & 0.36                                                & 0.47    & 0.38    \\ \cline{2-12} 
                           & \multirow{5}{*}{P8-E5} & \multicolumn{1}{c|}{AR}                                    & 1.186                                                 & 0.864   & \multicolumn{1}{c|}{0.997}   & 0.672   & 0.138   & 0.294   & 0.084                                               & 0.589   & 0.484   \\
                           &                      & \multicolumn{1}{c|}{SDR}                                   & 4.537                                                 & 4.814   & \multicolumn{1}{c|}{5.030}    & 4.769   & 3.472   & 3.825   & 2.046                                               & 4.779   & 4.268   \\
                           &                      & \multicolumn{1}{c|}{ATS}                                   & 690.51                                                & 701.13  & \multicolumn{1}{c|}{696.55}  & 723.58  & 749.65  & 741.73  & 764.18                                              & 699.74  & 733.95  \\
                           &                      & \multicolumn{1}{c|}{SDTS}                                  & 148.869                                               & 133.648 & \multicolumn{1}{c|}{148.770}  & 123.537 & 102.634 & 109.840  & 101.720                                              & 139.037 & 111.974 \\
                           &                      & \multicolumn{1}{c|}{SR}                                    & 0.52                                                  & 0.46    & \multicolumn{1}{c|}{0.48}    & 0.44    & 0.35    & 0.35    & 0.30                                                 & 0.46    & 0.37    \\ \bottomrule \toprule
\multirow{15}{*}{4×5}      & \multirow{5}{*}{P6-E3} & \multicolumn{1}{c|}{AR}                                    & -0.176                                                & -0.874  & \multicolumn{1}{c|}{-0.628}  & -1.249  & -2.179  & -1.979  & -1.395                                              & -1.903  & -2.302  \\
                           &                      & \multicolumn{1}{c|}{SDR}                                   & 3.769                                                 & 4.573   & \multicolumn{1}{c|}{2.301}   & 5.329   & 1.484   & 3.722   & 5.428                                               & 4.121   & 4.381   \\
                           &                      & \multicolumn{1}{c|}{ATS}                                   & 726.55                                                & 738.63  & \multicolumn{1}{c|}{734.33}  & 742.07  & 748.62  & 745.99  & 739.82                                              & 743.47  & 738.68  \\
                           &                      & \multicolumn{1}{c|}{SDTS}                                  & 126.891                                               & 124.957 & \multicolumn{1}{c|}{130.540}  & 124.271 & 109.798 & 115.423 & 130.294                                             & 113.959 & 122.204 \\
                           &                      & \multicolumn{1}{c|}{SR}                                    & 0.39                                                  & 0.35    & \multicolumn{1}{c|}{0.38}    & 0.33    & 0.27    & 0.27    & 0.31                                                & 0.29    & 0.27    \\ \cline{2-12} 
                           & \multirow{5}{*}{P7-E4} & \multicolumn{1}{c|}{AR}                                    & -1.728                                                & -1.658  & \multicolumn{1}{c|}{-1.643}  & -1.842  & -2.191  & -1.916  & -1.863                                              & -2.100    & -2.388  \\
                           &                      & \multicolumn{1}{c|}{SDR}                                   & 3.894                                                 & 4.097   & \multicolumn{1}{c|}{3.672}   & 3.783   & 0.399   & 2.768   & 4.509                                               & 2.949   & 3.886   \\
                           &                      & \multicolumn{1}{c|}{ATS}                                   & 724.81                                                & 736.23  & \multicolumn{1}{c|}{731.06}  & 739.24  & 765.01  & 761.77  & 759.73                                              & 774.56  & 774.72  \\
                           &                      & \multicolumn{1}{c|}{SDTS}                                  & 130.391                                               & 118.947 & \multicolumn{1}{c|}{114.458} & 127.289 & 111.900   & 124.328 & 132.343                                             & 74.837  & 78.892  \\
                           &                      & \multicolumn{1}{c|}{SR}                                    & 0.32                                                  & 0.29    & \multicolumn{1}{c|}{0.30}     & 0.27    & 0.21    & 0.22    & 0.25                                                & 0.16    & 0.13    \\ \cline{2-12} 
                           & \multirow{5}{*}{P8-E5} & \multicolumn{1}{c|}{AR}                                    & -1.892                                                & -2.285  & \multicolumn{1}{c|}{-1.972}  & -2.314  & -2.785  & -2.735  & -2.187                                              & -2.026  & -3.725  \\
                           &                      & \multicolumn{1}{c|}{SDR}                                   & 3.075                                                 & 2.819   & \multicolumn{1}{c|}{3.012}   & 3.396   & 2.722   & 2.431   & 3.059                                               & 3.229   & 2.646   \\
                           &                      & \multicolumn{1}{c|}{ATS}                                   & 758.66                                                & 760.63  & \multicolumn{1}{c|}{765.71}  & 772.47  & 781.62  & 781.54  & 779.42                                              & 773.13  & 784.37  \\
                           &                      & \multicolumn{1}{c|}{SDTS}                                  & 79.871                                                & 85.183  & \multicolumn{1}{c|}{78.258}  & 83.856  & 23.915  & 37.835  & 35.683                                              & 76.244  & 12.699  \\
                           &                      & \multicolumn{1}{c|}{SR}                                    & 0.18                                                  & 0.16    & \multicolumn{1}{c|}{0.16}    & 0.15    & 0.11    & 0.11    & 0.12                                                & 0.15    & 0.09    \\ \bottomrule \toprule
\multirow{15}{*}{Real Map} & \multirow{5}{*}{P6-E3} & \multicolumn{1}{c|}{AR}                                    & -0.871                                                & -1.835  & \multicolumn{1}{c|}{-1.399}  & -2.136  & -2.903  & -2.264  & -2.115                                              & -1.950   & -2.567  \\
                           &                      & \multicolumn{1}{c|}{SDR}                                   & 1.194                                                 & 0.941   & \multicolumn{1}{c|}{1.394}   & 0.856   & 0.769   & 0.971   & 0.832                                               & 1.326   & 0.983   \\
                           &                      & \multicolumn{1}{c|}{ATS}                                   & 617.48                                                & 635.37  & \multicolumn{1}{c|}{624.46}  & 645.53  & 663.81  & 651.74  & 647.91                                              & 633.82  & 644.35  \\
                           &                      & \multicolumn{1}{c|}{SDTS}                                  & 197.582                                               & 206.494 & \multicolumn{1}{c|}{200.281} & 191.161 & 188.034 & 192.899 & 200.556                                             & 198.539 & 204.794 \\
                           &                      & \multicolumn{1}{c|}{SR}                                    & 0.68                                                  & 0.59    & \multicolumn{1}{c|}{0.62}    & 0.55    & 0.49    & 0.50     & 0.51                                                & 0.58    & 0.47    \\ \cline{2-12} 
                           & \multirow{5}{*}{P7-E4} & \multicolumn{1}{c|}{AR}                                    & -1.497                                                & -2.849  & \multicolumn{1}{c|}{-2.068}  & -3.097  & -3.773  & -3.318  & -3.019                                              & -2.971  & -3.416  \\
                           &                      & \multicolumn{1}{c|}{SDR}                                   & 0.835                                                 & 0.954   & \multicolumn{1}{c|}{1.261}   & 0.751   & 0.662   & 0.892   & 0.767                                               & 0.946   & 0.596   \\
                           &                      & \multicolumn{1}{c|}{ATS}                                   & 635.90                                                 & 657.62  & \multicolumn{1}{c|}{643.58}  & 665.49  & 674.75  & 660.97  & 668.32                                              & 677.15  & 685.13  \\
                           &                      & \multicolumn{1}{c|}{SDTS}                                  & 195.644                                               & 189.341 & \multicolumn{1}{c|}{201.894} & 181.518 & 183.612 & 208.259 & 180.703                                             & 186.439 & 126.547 \\
                           &                      & \multicolumn{1}{c|}{SR}                                    & 0.62                                                  & 0.48    & \multicolumn{1}{c|}{0.53}    & 0.45    & 0.41    & 0.47    & 0.43                                                & 0.39    & 0.30     \\ \cline{2-12} 
                           & \multirow{5}{*}{P8-E5} & \multicolumn{1}{c|}{AR}                                    & -2.629                                                & -3.326  & \multicolumn{1}{c|}{-2.989}  & -4.076  & -4.698  & -3.871  & -3.809                                              & -3.275  & -3.984  \\
                           &                      & \multicolumn{1}{c|}{SDR}                                   & 0.801                                                 & 0.592   & \multicolumn{1}{c|}{0.836}   & 0.410    & 0.470    & 0.837   & 0.966                                               & 0.714   & 0.241   \\
                           &                      & \multicolumn{1}{c|}{ATS}                                   & 662.33                                                & 684.54  & \multicolumn{1}{c|}{671.44}  & 701.88  & 694.29  & 684.16  & 678.16                                              & 684.61  & 707.06  \\
                           &                      & \multicolumn{1}{c|}{SDTS}                                  & 153.710                                                & 175.043 & \multicolumn{1}{c|}{160.515} & 150.537 & 139.478 & 179.114 & 178.559                                             & 180.397 & 155.917 \\
                           &                      & \multicolumn{1}{c|}{SR}                                    & 0.56                                                  & 0.41    & \multicolumn{1}{c|}{0.48}    & 0.36    & 0.34    & 0.41    & 0.37                                                & 0.38    & 0.27    \\ \bottomrule
\end{tabular}
\end{center}
\label{tabel_result}
\end{table*}

\begin{figure*}[]
\centering
\centerline{\includegraphics[width=\textwidth,height=5.6cm]{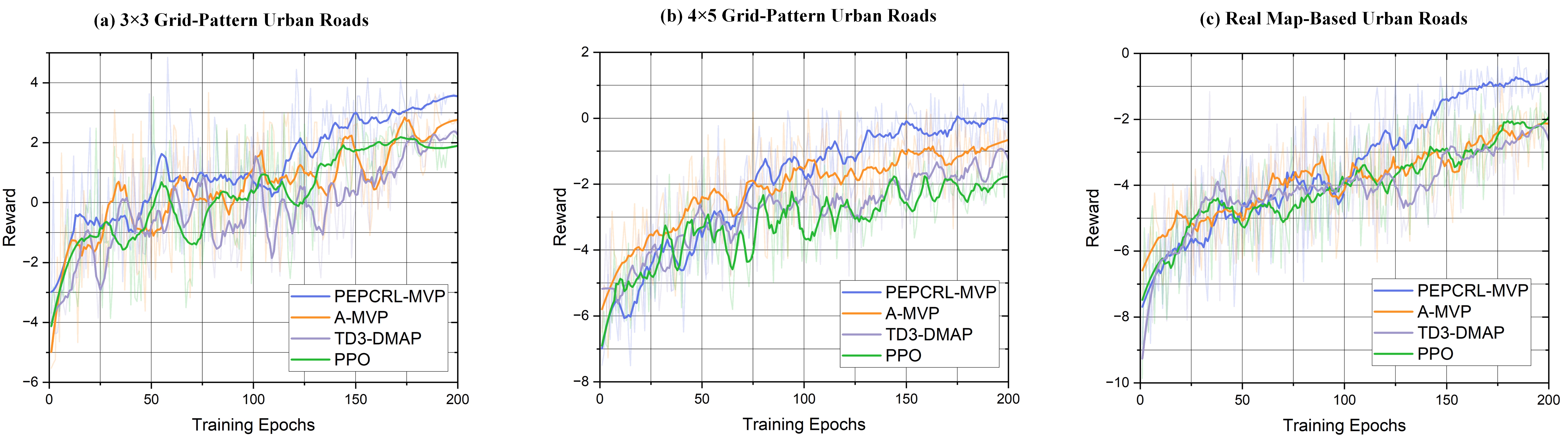}}
\caption{Training reward of P6-E3 in different scenes.}
\label{fig4}
\end{figure*}

\subsection{Ablation Experiments}
We conduct 100 tests on every model and measure the pursuit performance in terms of five metrics, which are average reward (AR), the standard deviation of reward (SDR), average time steps (ATS), the standard deviation of time steps (SDTS), and the pursuing success rate (SR).  Ablation experiments are designed to investigate the effect of the proposed prioritized experience selection and progression cognition modules in PEPCRL-MVP. The ablation experiment results are shown in columns 4 to 6 in TABLE \ref{tabel_result}. The results of A-MVP correspond to a DQN-based path planning with progression cognition without prioritized experience selection, and the results of B-MVP correspond to a DQN-based path planning equipped with prioritized experience selection without attention-based progression cognition.

PEPCRL-MVP shows the best performance in scenes with different difficulty levels under the same urban traffic road structure. In the real map-based urban road, compared with A-MVP, the AR of PEPCRL-MVP increases $52.53\%$, $47.46\%$, and $20.96\%$, respectively, and the ATS of PEPCRL-MVP decreases $2.82\%$, $3.31\%$, and $3.24\%$ in P6-E3, P7-E4, and P8-E5 difficulty levels, respectively. Furthermore, the SDR and SDTS of PEPCRL-MVP are comparable to those of A-MVP. These results reveal that prioritized experience selection can effectively promote cooperation among pursuing vehicles and improve pursuing performance. 

Given the pursuing difficulty level, PEPCRL-MVP also substantially outperforms the other methods under different urban traffic road scenes. Taking the P6-E3 as an example, the SDTS of PEPCRL-MVP is $7.55\%$, $2.80\%$ and $1.35\%$ lower than that of B-MVP under $3 \times 3$, $4 \times 5$, and real map-based scenes, respectively. The results show the proposed method has excellent robustness. The SDR is $10.72\%$ lower than that of B-MVP on average in the P7-E4 difficulty level. These results can be explained by the fact that attention-based progression cognition can considerably enhance the stability of pursuing vehicles' performance. Moreover, the proposed PEPCRL-MVP has a better generalization and pursuing validity than A-MVP and B-MVP. It is evident that in different scenes, whether the urban road structure or the pursuing difficulty level is different, PEPCRL-MVP has the greatest SR. Concretely, the SR of PEPCRL-MVP is $19.67\%$, $11.92\%$ higher than that of A-MVP and B-MVP on average, respectively.

In addition, to investigate the impact of the prioritized experience selection on the PEPCRL-MVP's training convergence, we present the average reward with the P6-E3 setting in Fig. \ref{fig4}. Fig. \ref{fig4} (a) shows the average reward with training epochs under the $3 \times 3$ grid pattern. It can be noticed that compared with A-MVP, PEPCRL-MVP has a smaller fluctuation and a more stable convergence in the late stage of training. For the $4\times5$ grid pattern, as shown in Fig. \ref{fig4} (b), although the average reward of A-MVP grows fast in the early training stage, PEPCRL-MVP has a faster convergence speed and higher convergence trend in the late training stage. In Fig. \ref{fig4} (c), under the real map-based urban road, PEPCRL-MVP shows an obviously stronger convergence. According to the results in Fig \ref{fig4}, it can be observed that the prioritized experience selection greatly facilitates the convergence of agent learning, and helps improve the pursuing performance.

The above results obtained from the ablation experiments demonstrate that the prioritized experience selection network can select appropriate training set for each agent, which overcomes the problem of agent differentiation in the existing MVP approaches. It can effectively promote the convergence of agent learning, and enhance cooperation among agents. Furthermore, the progression cognition module can decide appropriate targets for each pursuing vehicle according to the real-time traffic situation and MVP task, consequently improving pursuing efficiency and system stability.

\subsection{Comparison with Other Methods}
We compare the PEPCRL-MVP to the other state-of-the-art RL approaches for MVP, including DQN, DDPG, MADDPG, Twin Delayed Deep Deterministic policy gradient-Decentralized Multi-Agent Pursuit (TD3-DMAP) \cite{mvp3}, Proximal Policy Optimization (PPO), and Transformer-based Time and Team RL for Observation-constrained MVP (T$^3$OMVP) \cite{mvp6}. The comparison results are presented in the columns beginning from column 7 in TABLE \ref{tabel_result}. 

According to the comparison results, for any given pursuing difficulty level, PEPCRL-MVP shows remarkable performance improvement under all the traffic road scenes. For the setting of P8-E5, the AR of PEPCRL-MVP is improved by $43.41\%$ and $42.5\%$ on average, respectively, compared with DQN and PPO under the $3 \times 3$, $4 \times 5$ and real map-based road structures, which are the top two performances of all comparison methods in general. 
And the ATS of PEPCRL-MVP decreases by $4.18\%$ on average compared with other methods in the P8-E5 difficulty level under the real map-based urban road. The results show that the proposed PEPCRL-MVP approach can highly improve pursuing efficiency and have excellent adaptability to different road scenes and traffic situations.

Under the same urban traffic road scenes, PEPCRL-MVP shows competitive robustness and pursuing effectiveness at different pursuing difficulty levels. Under the $3\times3$ grid pattern urban roads, as the pursuing difficulty level increases, the ATS of PEPCRL-MVP is $6.46\%$, $1.73\%$, and $1.32\%$ lower than that of PPO which has the second-best performance, in the P6-E3, P7-E4, and P8-E5 difficulty levels respectively. Comprehensively, we evaluate the SR metric for all methods to further compare PEPCRL-MVP to other methods. The SR of PEPCRL-MVP is $51.09\%$ higher than that of other methods on average under the $4 \times 5$ scene, and $47.53\%$ higher than that of other methods on average for all scenes. These results indicate that PEPCRL-MVP greatly improves pursuing efficiency and has stronger robustness.

It is noted that there is no significant SR and SDTS performance improvement by PEPCRL-MVP. This can be explained by the fact that we set the maximum time steps to 800, which leads to a higher standard deviation for the methods with better performance in the 100 tests. Although the SDR and SDTS do not obtain great results, there is a considerable increase in AS, ATS, and SR for the PEPCRL-MVP approach. Specifically, in the P7-E4 difficulty level under the real map-based simulation environment, even if the SDR and SDTS of PEPCRL-MVP are $11.16\%$ and $7.78\%$ higher than those of DQN, the AR of PEPCRL-MVP is greatly improved, increasing by $51.66\%$. The SDR and SDTS of PEPCRL-MVP are respectively $11.54\%$ and $3.37\%$ lower than those of PPO, which has the second-best performance. At the same time, the AR increases by $40.81\%$, and the ATS decreases by $6.46\%$ in the P6-E3 difficulty level under the $3 \times 3$ road scene. These results demonstrate that PEPCRL-MVP can achieve great improvements both in algorithm stability and pursuing efficiency in simple scenes, and sacrifices some stability to obtain higher pursuing efficiency and better average performance in some complex scenes.

\begin{figure*}[t]

\centering
\centerline{\includegraphics[width=\textwidth]{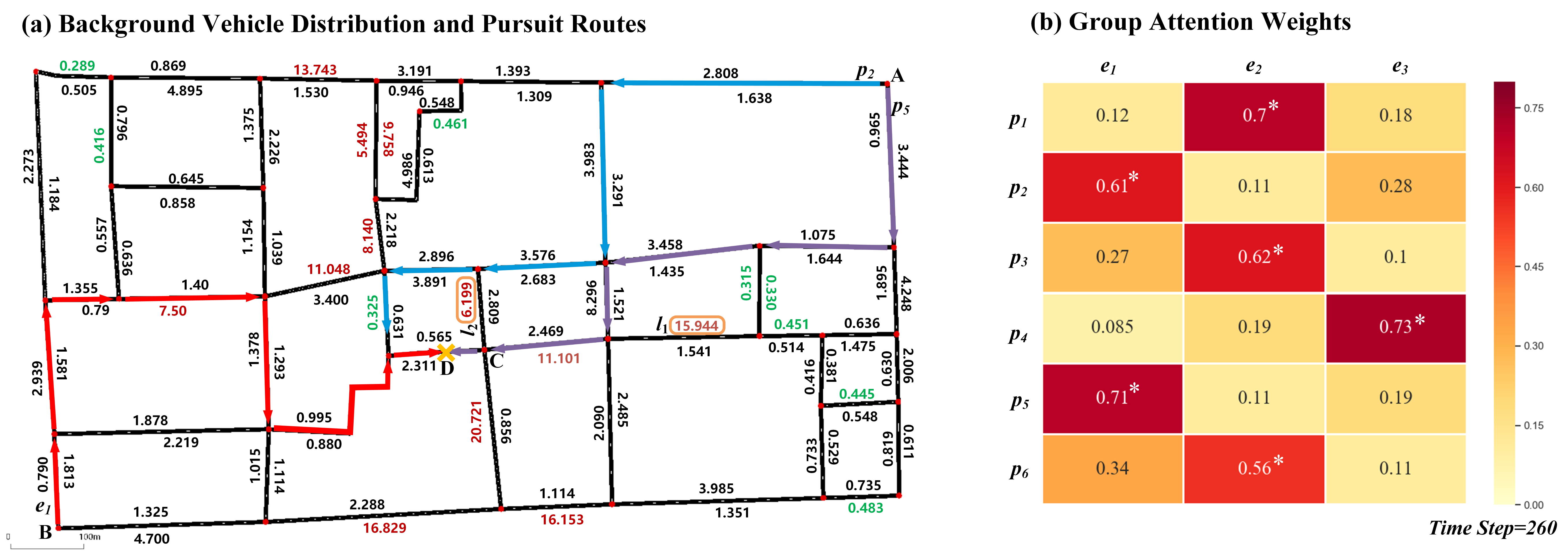}}
\caption{Case study of P6-E3 in real map-based scene.}
\label{fig6}
\end{figure*}

Fig. \ref{fig4} describes the convergence curve of average reward with training epochs of the P6-E3 under different road structures. 
In Fig. \ref{fig4} (a), both TD3-DMAP and PPO have large fluctuations in the late stage of training under the $3\times 3$ road structure scene, while TD3-DMAP shows a superior growth trend and stable convergence. Fig. \ref{fig4} (b) depicts the average reward under the $4\times5$ urban road scene, presenting that the PEPCRL-MVP has advantageous performances in both convergence rate and convergence stability compared with TD3-DMAP and PPO. For the real map-based urban road scene, as shown in Fig. \ref{fig4} (c), compared with other methods, PERL has a better convergence trend and higher reward. In conclusion, Fig. \ref{fig4} illustrates that compared with TD3-DMAP and PPO, which belong to the superior performance in all comparison methods, PEPCRL-MVP makes the competitive convergence trend and stability, demonstrating its superiority and effectiveness.

\subsection{Case Study}

In this section, we analyze the PEPCRL-MVP pursuing processes in a case study with a real map-based scene in detail. Representative results are shown in Fig. \ref{fig6}. Fig. \ref{fig6} (a) presents the distribution of background vehicles and the pursuing routes, where, the number marked next to the lane represents the average number of background vehicles per time step in this lane during the pursuit. If the average number of vehicles in a lane is greater than 10, it means that the lane is congested. Also, Fig. \ref{fig6} (a) shows the routes of pursuing vehicles $p_2$ and $p_5$. 
Following the group attention weights, $p_2$ and $p_5$ form a group to capture $e_1$ from $A$. 
From the routes of $p_2$ and $p_5$, it can be seen that they predict the trajectory of the target evading vehicle and collaboratively pursue and intercept $e_1$. It is worth noting that $p_5$ plans the path to $C$ while avoiding the congested lane $l_1$ and lane $l_2$. Finally, $p_5$ catches $e_1$ at $D$. This pursuit process shows the efficient cooperation of $p_2$ and $p_5$. 

Fig. \ref{fig6} (b) shows the group attention weights in 260 time steps during the pursuit. It can be observed that $p_2$ and $p_5$ both focus their attention on $e_1$. Moreover, each pursuing vehicle has its own target evading vehicle and each evading vehicle may be pursued by one or more pursuing vehicles. 
It indicates that the progression cognition module can select suitable targets for pursuing vehicles according to the traffic situation and the locations of evading vehicles. The PEPCRL-MVP can achieve efficient and effective collaborative multi-vehicle pursuit.

\begin{figure}[h]
\centering
\centerline{\includegraphics[width=0.8\columnwidth]{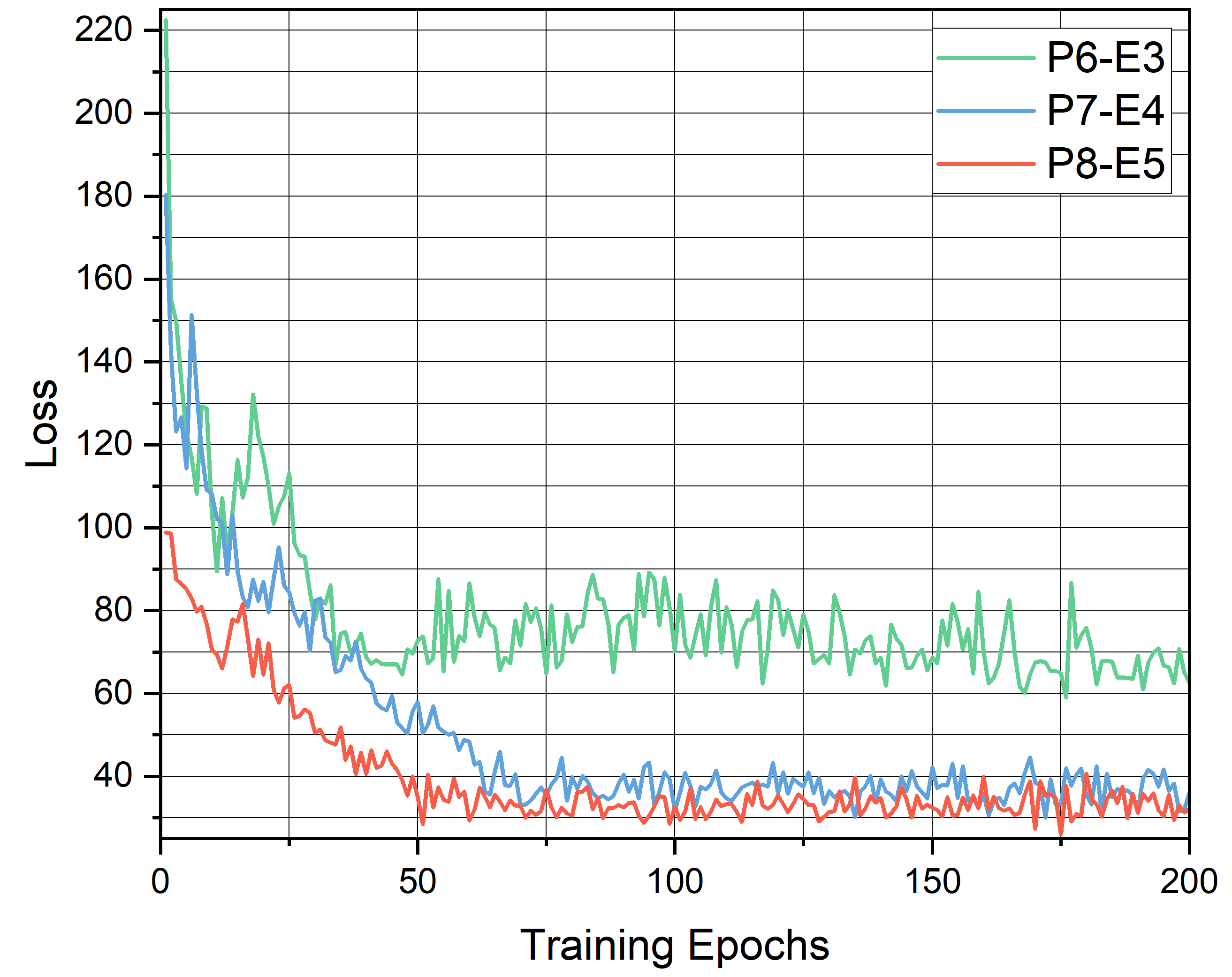}}
\caption{Training loss of prioritization network in the real map-based scene.}
\label{fig5}
\end{figure}

To investigate the impact of the prioritization network, we show the training loss of the prioritization network in the real map-based scene in Fig. \ref{fig5}. The prioritization networks with different numbers of pursuing and evading vehicles all converge in 100 training epochs. The training of the prioritization network with a smaller number of agents converges more easily, but the mean square error between the network output and reward gain is larger. It is clear from Fig. \ref{fig5} that as the number of agents increases, the loss of the prioritization network decreases. This suggests that extensive experience collection greatly contributes to prioritized network performance. It also demonstrates that the prioritization network can effectively evaluate the global experience pool, thus facilitating the learning and collaboration of multiple agents.

\section{Conclusion} \label{conclu}
The emerging MARL technology is promising for multi-vehicle pursuit applications. However, the mission and safety-critical MVP tasks present great challenges, especially for the chasing of multiple target vehicles. While there are existing MARL algorithms proposed for MVP, they usually applied centralized training with randomly selected experience samples and did not adapt well to dynamically changing traffic situations. To address the problems in the existing MVP algorithm, in this paper we proposed a novel MVP approach (called PEPCRL-MVP) to improve MARL learning, collaboration, and MVP performance in dynamic urban traffic scenes. There are two major new components included in PEPCRL-MVP, a prioritization network and an attention-based progression cognition module. The prioritization network was introduced to effectively select training experience samples and increase diversity for the optimization and behavior of MARL, which improved agent collaboration and extensive exploration of experience. The progression cognition module was introduced to extract key traffic features from the sensor data and support the pursuing vehicles to adaptive adjust their target evading vehicles and path planning according to the real-time traffic situations. 
A simulator was developed for evaluation of the proposed PEPCRL-MVP approach and comparison with existing ones. Extensive experiments were conducted over urban roads in an area inside the second ring road of Beijing on PEPCRL-MVP and several approaches. Experiment results demonstrate that PEPCRL-MVP significantly outperform the other methods for all the investigated road scenes in terms of performance metrics including pursuing success rate and average rewards. The results also demonstrate the effectiveness of the proposed two components. 
Jointly they largely improve collaboration and traffic awareness, leading to improved MVP performance. In the future, we will investigate the impact of additional factors in MVP, such as pedestrians, social activities, and communication delay, on the design and analysis of MVP approaches.


\section*{Acknowledgments}
This work is supported by the National Natural Science Foundation of China (Grant No. 62176024), the National Key R\&D Program of China (2022ZD01161, 2022YFB2503202), Beijing Municipal Science \& Technology Commission (Grant No. Z181100001018035) and Engineering Research Center of Information Networks, Ministry of Education.
\bibliographystyle{IEEEtran}
\bibliography{reference_TIV}

\vfill

\end{document}